\documentclass[runningheads,a4paper]{llncs}

\usepackage[pagebackref=false,breaklinks=true,colorlinks,bookmarks=false,urlcolor=blue,linkcolor=blue,citecolor=blue,pdftitle={Diversified Late Acceptance Search},pdfauthor={Majid Namazi, Conrad Sanderson, M.A. Hakim Newton, M.M.A. Polash, Abdul Sattar}]{hyperref}
\usepackage{amssymb}
\usepackage{graphicx}
\usepackage{amssymb}
\usepackage{adjustbox}
\usepackage{url}

\usepackage{color}
\usepackage{soul}
\definecolor{lightgray}{gray}{0.85}
\sethlcolor{lightgray}

\usepackage{times}  % changes the default serif font
\usepackage[scaled=0.9]{inconsolata}  % changes the default typewriter font

\begin{document}

\mainmatter  

\title{Diversified Late Acceptance Search}
%\titlerunning{Diversified Late Acceptance Search}
\titlerunning{~}

\author
  {
  Majid Namazi$~^{1,2}$,
  Conrad Sanderson$~^{2,3}$,
  M.A.~Hakim Newton$~^1$,\\
  M.M.A.~Polash$~^1$,
  Abdul Sattar$~^{1}$
  }

\institute
  {
  $^1$~Griffith University, Australia\\
  $^2$~Data61, CSIRO, Australia\\
  $^3$~University of Queensland, Australia
  ~\textcolor{white}{\thanks{\scriptsize {\bf Published in:} Lecture Notes in Computer Science (LNCS), Vol.~11320, pp.~299-311, 2018.\protect\\ \href{https://doi.org/10.1007/978-3-030-03991-2_29}{\tt https://doi.org/10.1007/978-3-030-03991-2\_29}}}
  }

%\authorrunning{Namazi, Sanderson, Newton, Polash, Sattar}
\authorrunning{~}

\maketitle

\vspace{-4ex}
\begin{abstract}

The well-known Late Acceptance Hill Climbing (LAHC) search aims
to overcome the main downside of traditional Hill Climbing (HC) search,
which is often quickly trapped in a local optimum
due to strictly accepting only non-worsening moves within each iteration.
In contrast, LAHC also accepts worsening moves,
by keeping a circular array of fitness values of previously visited solutions
and comparing the fitness values of candidate solutions against the least recent element in the array.
While this straightforward strategy has proven effective,
there are nevertheless situations where LAHC can unfortunately
behave in a similar manner to HC.
For example, when a new local optimum is found, often the same fitness value is stored many times in the array.
To address this shortcoming,
we propose new acceptance and replacement strategies
to take into account worsening, improving, and sideways movement scenarios
with the aim to improve the diversity of values in the array.
Compared to LAHC,
the proposed Diversified Late Acceptance Search approach is shown
to lead to better quality solutions that are obtained with a lower number of iterations
on benchmark Travelling Salesman Problems and Quadratic Assignment Problems.

% \vspace{-1.5ex}
% \keywords{Local Search, Late Acceptance, Diversification.}
\vspace{-2.5ex}

\end{abstract}

\pagestyle{empty}

\section{Introduction}
\vspace{-1.5ex}

Local search algorithms are typically efficient and scalable approaches to solve large instances of real world optimisation problems~\cite{Bhardwaj_2018,hoos2004stochasticIvD}.
Such algorithms use the following overall approach: starting from an {\em initial solution}, iteratively move from one solution to another,
with the aim to eventually arrive at a good solution.
The initial solution is often generated randomly or by using a specialised method.
Then, in each iteration, a {\em candidate solution} is obtained by modifying the {\em current solution} using a {\em perturbation method}.
If the candidate solution in a given iteration satisfies a given {\em acceptance criterion}, it is used as the starting point for the next iteration.
Otherwise, the current solution in the given iteration becomes the starting point for the next iteration.
The traditional Hill Climbing (HC) approach is a local search method that strictly uses a greedy strategy as its acceptance criterion~\cite{appleby1961techniquesHC}.
HC accepts the candidate solution only if its fitness value is better
(smaller in minimisation problems and larger in maximisation problems) than that of the current solution.
This greedy strategy typically leads the search to quickly being trapped in a local optimum.

An important challenge in designing a local search algorithm is to find a good balance between 
interleaving {\em diversification} and {\em intensification} phases during search~\cite{hoos2004stochasticIvD}.
{\em Diversification} means exploring the solution space as widely as possible,
with the intent of ideally finding a globally optimum solution.
In contrast, {\em intensification} means improving the current solution in order to converge to the best local solution as quickly as possible.
The perturbation method as well as the acceptance criterion need to take this balancing issue into account.
As HC does not explore solutions that are worse than the current solution in each iteration,
HC uses a very high level of intensification at the cost of very low level of diversification.
Overall, the HC algorithm converges quickly to a local optimum, but the quality of its solutions is often not high~\cite{Burke08alateLAHC08,BURKE201770LAHC17}. 
Diversification strategies are hence necessary to provide better solutions.

There are well-studied acceptance criteria that, with the aim to avoid or escape local optima, also accept worsening moves,
rather than simply accepting only better candidate solutions.
Simulated Annealing (SA) \cite{kirkpatrick1983optimizationSA} uses a stochastic 
acceptance criterion, where worsening moves are accepted
with a probability based on the difference in the fitness values of the current solution and the candidate solution,
with the probability exponentially diminishing over time.
Threshold Acceptance (TA)~\cite{dueck1990thresholdTA} is a deterministic acceptance criterion,
which accepts worsening moves if the difference in the fitness values of the current and the candidate solution is below a given threshold.
The Great Deluge Algorithm  (GDA)~\cite{dueck1993newGDA,mcmullan2007extended,obit2009NonLinVarSch}
accepts worsening moves if the fitness value of the candidate solution 
is below a given level. Each of the above acceptance criteria has a parameter whose {\em initial value}
and a {\em variation schedule} must be defined beforehand.
Unfortunately, obtaining a suitable initial value and variation schedule is difficult to achieve,
and is often problem domain dependent and/or problem instance dependent~\cite{burke2004HighInitVal,BURKE201770LAHC17,mcmullan2007extended}.
This can make practical use of SA, TA and GDA quite finicky.

In contrast to the above approaches, Late Acceptance Hill Climbing (LAHC) search~\cite{Burke08alateLAHC08,BURKE201770LAHC17}
is a relatively straightforward technique which deterministically accepts worsening moves
and has no complicated parameters.
An array with a predefined length stores the fitness values of previously visited solutions.
Fitness values of candidate solutions are compared against the least recent element in the array.
Since the fitness values from previous iterations can be worse than that of the current solution,
a candidate solution that is worse than the current solution can be accepted.
As the search progresses, the array is deterministically updated with fitness values of new solutions.
The use of the fitness array thus brings about search diversity.
The larger the length of the array, the better the diversity level.
Overall, LAHC exhibits better diversification in terms of the explored solutions
and provides solutions which typically have higher quality than HC~\cite{Burke08alateLAHC08,BURKE201770LAHC17}.
Moreover, LAHC has been successful in several optimisation competitions~\cite{afsar2016machineCompet,wauters2015winningCompet},
and has been used in real world applications~\cite{optaplannerApp}.

Despite the promising aspects of LAHC, in this work we observe that there are situations
where LAHC can unfortunately behave in a similar manner to HC,
even when using a large fitness array.
For example, when the same fitness value is stored many times in the array,
particularly when a new local optimum is found.
In this case, the fitness values in the array are iteratively replaced with the new local optimum fitness value,
thereby reducing diversity.

To address the above shortcoming, we propose a new search approach termed Diversified Late Acceptance Search (DLAS).
With the aim to improve the overall diversity of the search,
the approach uses:
{\bf (i)} a new acceptance strategy which increases diversity of the accepted solutions,
and
{\bf (ii)} a new replacement strategy to improve the diversity of the values in the fitness array by taking worsening, improving, and sideways movement scenarios into account.

Section~\ref{secLateAccept} overviews the LAHC algorithm and discusses its problems.
Section~\ref{secDiverseLateAccept} presents the proposed DLAS algorithm.
Section~\ref{secExperiments} provides comparative evaluations
on benchmark Travelling Salesman Problems (TSPs)
and Quadratic Assignment Problems (QAPs).
The main findings are summarised in Section~\ref{secConclusions}.

\section{Late Acceptance Hill Climbing}
\label{secLateAccept}
\vspace{-1ex}

Local search algorithms start from an {\em initial solution} {\small $S_0$}.
The current solution {\small $S_k$} in each iteration {\small $k$} is then modified by a given perturbation method {\small $M$} to generate a new candidate solution {\small $S'_k = M(S_k)$}.
Next, using a given {\em acceptance criterion} {\small $\mathcal{A}$},
the candidate solution {\small $S'_k$} is either accepted or rejected,
meaning either {\small $S_{k+1} = S'_k$} if {\small $\mathcal{A}(k) = \textsf{true}$},
or {\small $S_{k+1} = S_k$} if {\small $\mathcal{A}(k) = \textsf{false}$}.
Assume {\small $F_k$} and {\small $F'_k$} denote the fitness values of solutions {\small $S_k$} and {\small $S'_k$}, respectively.
For convenience, we assume {\em minimisation problems}, where one solution is {\em better} than the other if fitness value of the former is less than that of the latter.
In HC, {\small $\mathcal{A}(k) = \textsf{true}$} iff {\small $F'_k \leq F_k$}, and so {\small $F_k \geq F_{k+1}$} for all {\small $k \geq 0$}.
Hence HC accepts only non-worsening moves, ie., sideways moves or improving moves.

The most recent version of LAHC \cite{BURKE201770LAHC17} accepts candidate solution {\small $S'_k$}
if its fitness value {\small $F'_k$} is better than or equal to the fitness value {\small $F_k$} of the current solution {\small $S_k$}, as in HC.
Furthermore, for a given {\em history length} {\small $L$}, candidate solution {\small $S'_k$} is accepted
if its fitness value {\small $F'_k$} is better than the fitness value {\small $F_{k-L}$} of the then current solution {\small $S_{k-L}$} at iteration {\small $k-L \geq 0$}.
In other words, {\small $\mathcal{A}(k) = F'_k \leq F_k$} or {\small $F'_k < F_{k-L}$} for {\small $k \geq L$}.  
Since {\small $F_{k-L}$} is usually (not always as in HC) greater than {\small $F_k$}, 
the candidate solution {\small $S'_k$} can be accepted at iteration {\small $k \geq L$}, even if {\small $F'_k > F_k$}.
LAHC thus accepts worsening moves like TA and GDA and thereby aims to avoid or escape from local minima. 
Overall, LAHC exhibits better diversification level with a larger {\small $L$}~\cite{Bazargani2017pLAHC,BURKE201770LAHC17},
as this allows comparison with further earlier solutions which are most likely further worse as well.

\figurename~\ref{algoLAHC} shows the pseudo code for LAHC.
To achieve memory efficiency, a circular {\em fitness array} {\small $\Phi$} of size {\small $L$} stores fitness values of previous {\small $L$} solutions.
Initially all values in {\small $\Phi$} are set to the initial {\small $F$}, ie.,  {\small $F_0$} (line~4).
Note that {\small $F$}, {\small $F'$}, {\small $S$} and {\small $S'$} at each iteration {\small $k$} in \figurename~\ref{algoLAHC}
correspond to {\small $F_k$}, {\small $F'_k$}, {\small $S_k$} and {\small $S'_k$}, respectively.
A~candidate solution $S'$ is accepted if {\small $F' \leq F$} or {\small $F' <\Phi[l]$}
where {\small $l=k \textsf{ mod } L$} (lines 9-10).
The value in {\small $\Phi[l]$} is replaced by {\small $F$} whenever {\small $F < \phi[l]$} (lines 13-14).

\begin{figure}[!tb]
\begin{minipage}[t]{0.49\textwidth}
\centering
\scalebox{0.95}{
\begin{minipage}{1\textwidth}
\sf
\begin{tabbing}
AB\=A\=A\=A\=A\=A\=\kill
~1  \> {\bf proc LAHC}\\
~2  \>\>Initialise curr solution $S$, compute $F$\\
~3  \>\>Specify length $L$ for fitness array $\Phi$\\
~4  \>\>{\bf forall} $l \in [0, ‌L)$, $\Phi[l] \leftarrow F$\\
~5  \>\> $k \leftarrow 0, S_* \leftarrow S, F_* \leftarrow  F$ ~~~~~//~best $S_*$\\
~6  \>\> {\bf while} {\sf termination-criteria} ~~//~iter $k$\\
~7  \>\>\> $S' \leftarrow M(S)$, compute $F'$ //~perturb\\
~8  \>\>\> $l \leftarrow k \textsf{ mod } L$\\
~9  \>\>\> {\bf if} $F' \leq F$ {\bf or} $F' < \Phi[l]$  \\ 
10  \>\>\>\> $S \leftarrow S', F \leftarrow F'$  ~~~~~~~~~~~//~accept\\
11  \>\>\>\> {\bf if} $F < F_*$ \\
12  \>\>\>\>\>$S_* \leftarrow S, F_* \leftarrow F$ ~~~~~//~new best\\
13  \>\>\> {\bf if} $F < \Phi[l]$ \\
14  \>\>\>\>$\Phi[l] \leftarrow F$ ~~~~~~~~~~~~~~~//~replace in $\Phi$\\
15  \>\>\> $k \leftarrow k + 1$\\
16  \>\>return $S_*, F_*$
\end{tabbing}
\end{minipage}
}
\caption{Late Acceptance Hill Climbing (LAHC) algorithm, adapted from~\cite{BURKE201770LAHC17}.}
\label{algoLAHC}
\end{minipage}
\hfill
\begin{minipage}[t]{0.49\textwidth}
\centering
\scalebox{0.95}{
\begin{minipage}{1\textwidth}
\sf
\begin{tabbing}
AB\=A\=AB\=A\=A\=A\=\kill
~1  \> {\bf proc DLAS}\\
~2  \>\>Initialise curr solution $S$, compute $F$\\
~3  \>\>Specify length $L$ for fitness array $\Phi$\\
~4  \>\>{\bf forall} $l \in [0, ‌L)$, $\Phi[l] \leftarrow F$\\
~5 \>\> $\Phi_{\textrm{max}}\leftarrow F$, $N\leftarrow L$\\
~6  \>\> $k \leftarrow 0, S_* \leftarrow S, F_* \leftarrow F$ ~~~~~~~// best $S_*$\\
~7  \>\> {\bf while} {\sf termination-criteria}~~~~~// iter $k$\\
~8 \>\>\> $F^-\leftarrow F$ ~~~~~~~~~~~~~~~~~~~~~~~~// previous\\
~9  \>\>\> $S' \leftarrow M(S)$, compute $F'$ ~// perturb\\
10  \>\>\> $l \leftarrow k \textsf{ mod } L$\\
11  \>\>\> {\bf if} $F' = F$ {\bf or} $F' < \Phi_{\textrm{max}}$\\ 
12  \>\>\>\> $S \leftarrow S', F \leftarrow F'$ ~~~~~~~~~~~~~// accept\\
13  \>\>\>\> {\bf if} $F < F_*$  \\
14  \>\>\>\>\> $S_* \leftarrow S, F_* \leftarrow F$ ~~~~~~// new best\\
15  \>\>\> {\bf if} $F > \Phi[l]$\\
16  \>\>\>\> $\Phi[l] \leftarrow F$ ~~~~~~~~~~~~~~~~// replace in $\Phi$\\
17  \>\>\> {\bf else} {\bf if} $F < \Phi[l]$ {\bf and} $F < F^-$\\
18  \>\>\>\> {\bf if} $\Phi[l]=\Phi_{\textrm{max}}$\\
19 \>\>\>\>\> $N\leftarrow N-1$ ~~~~~~~~~~~// decrement\\
20  \>\>\>\> $\Phi[l] \leftarrow F$ ~~~~~~~~~~~~~~~~// replace in $\Phi$\\
21  \>\>\>\> {\bf if} $N=0$\\
22  \>\>\>\>\> compute $\Phi_{\textrm{max}}$,$N$ ~// recompute\\
23  \>\>\> $k \leftarrow k + 1$\\
24  \>\>return $S_*, F_*$
\end{tabbing}
\end{minipage}
}
\caption{Proposed Diversified Late Acceptance Search (DLAS).}
\label{algoDLAS}
\end{minipage}
\end{figure}

\vspace{-1ex}
\subsection{Problems with LAHC}
\label{secProblems}
\vspace{-1ex}

\begin{figure}[!t]
\begin{minipage}[t]{1\textwidth}
  \centering
  \begin{minipage}[t]{0.48\textwidth}
    \centering
    \includegraphics[width=1\textwidth]{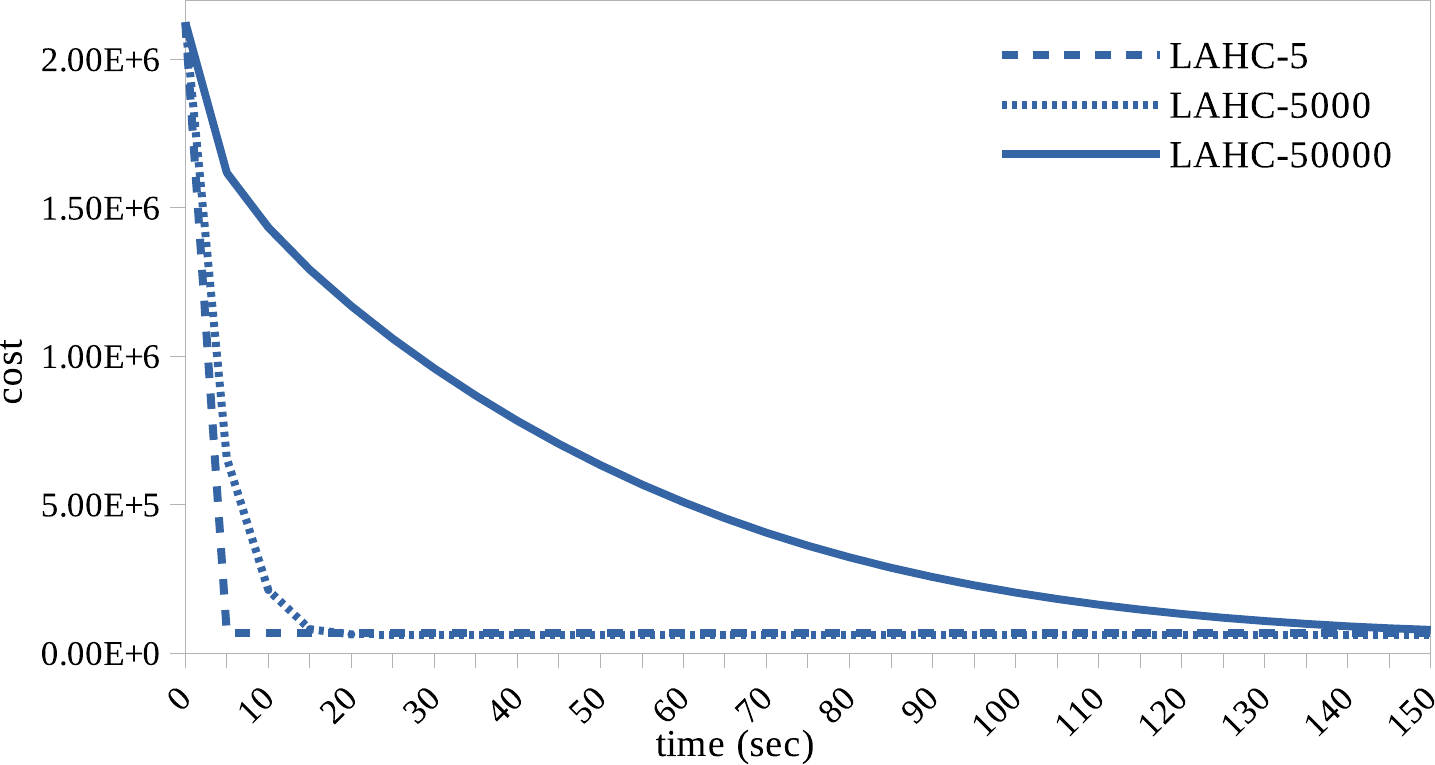}
    \vspace{-5ex}
    \caption
      {
      Search progress for the first 150 seconds while solving the benchmark U1817 TSP instance via LAHC with \mbox{$L$ $\in$ \{5, 5000, 50000\}}.
      Further progress is shown in Fig.~\ref{U1817Exp}.
      (To aid clarity, results for DLAS are not shown as they effectively cover LAHC with $L$=$5$ at the given scale.)
      }
    \label{U1817Fig}
    \vfill
  \end{minipage}
  \hfill
  \begin{minipage}[t]{0.48\textwidth}
    \centering
    \includegraphics[width=1\textwidth]{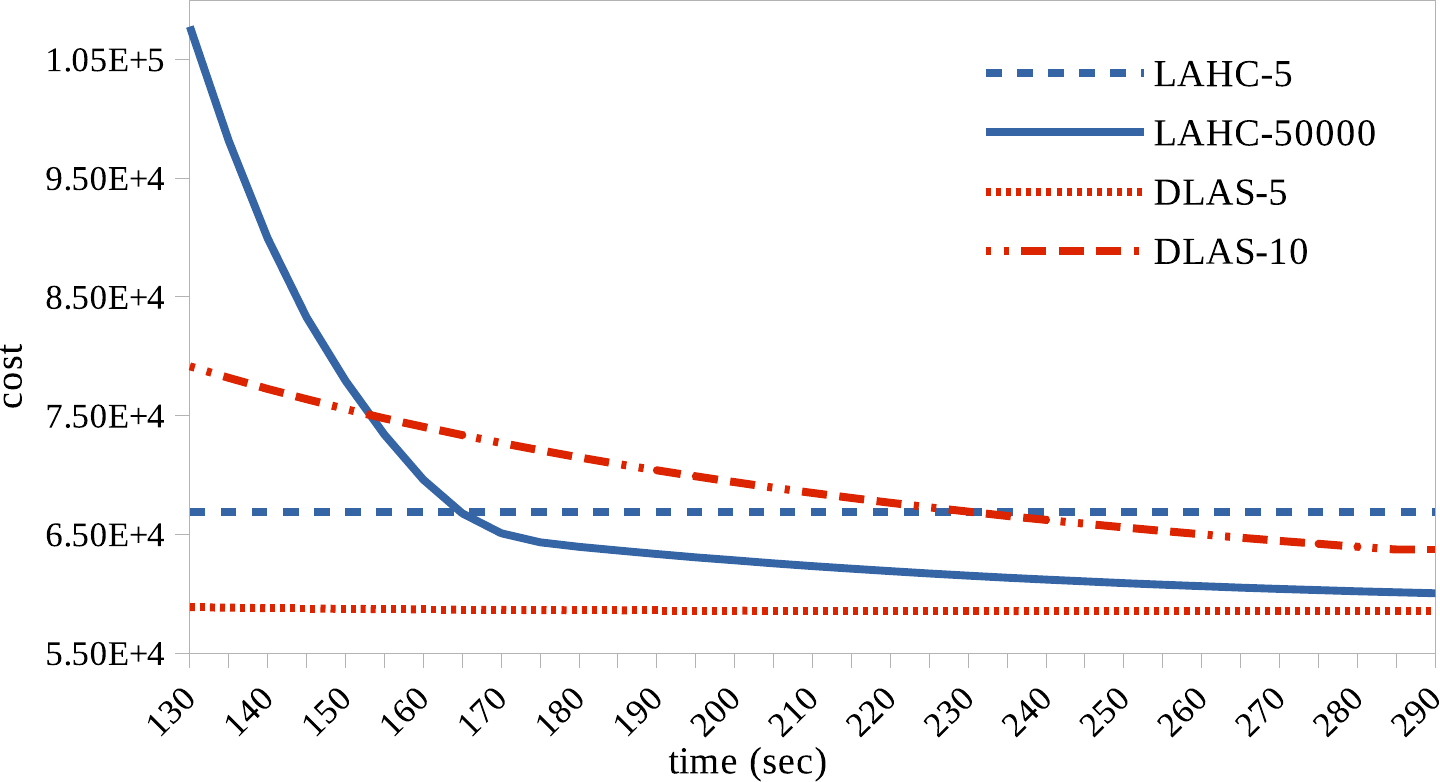}
    \vspace{-5ex}
    \caption
      {
      Search progress of LAHC and DLAS with various $L$ values in later iterations of solving the U1817 instance.
      LAHC with $L$=$5$ converges quicker than LAHC with $L$=$50000$, but obtains a worse solution.
      DLAS with $L$=$5$ obtains a better solution than LAHC.
      Furthermore, DLAS with $L$=$5$ converges much quicker than LAHC with $L$=$50000$.
      }
    \label{U1817Exp}
  \end{minipage}
\end{minipage}
\end{figure}

We have empirically observed that for some problems LAHC unfortunately behaves in a similar manner to HC and does not accept worsening moves.
Figs.~\ref{U1817Fig} and~\ref{U1817Exp} show typical search progress trend while solving the benchmark U1817 TSP instance (see Sec.~\ref{secExperiments} for TSP details).
A~similar pattern is seen in other benchmark instances.
For a small value of $L$, LAHC is quickly trapped in a local optimum,
leading to poor quality solutions.
Even using restart techniques may not help to obtain higher quality solutions~\cite{Bazargani2017pLAHC,BURKE201770LAHC17}.
For larger values of $L$ the search is less prone to trapping,
but this comes at the cost of slow convergence speed;
the solution quality can be poor if not enough time is allotted.
This characteristic of LAHC makes it less useful for applications in time-constrained systems where a high-quality solution must be found quickly.

The poor performance of LAHC is due to the following reasoning.
Consider the LAHC algorithm in \figurename~\ref{algoLAHC}. 
Assume that in a given iteration, all the values in the fitness array $\Phi$
are equal to the fitness value $F_*$ of a newly found best solution $S_*$, where $S_*$ is a hard-to-improve or a local optimum solution.
This happens when a new overall best solution $S_*$ with fitness value $F_*$ is found
and $F$ remains to be equal to $F_*$ for at least $L$ consecutive iterations.
In this case, no worsening moves with larger fitness values than $F_*$ will be accepted anymore,
and if $S_*$ is a local optimum then the search is trapped in that solution.
Clearly, this is the situation HC reaches when it is trapped in a local optimum.
In Sec.~\ref{secExperiments} we show that even when using a large value for $L$,
LAHC behaves like HC in solving many problems in a large proportion of the iterations.

\section{Proposed Diversified Late Acceptance Search}
\label{secDiverseLateAccept}
\vspace{-1ex}

We propose a new search approach that aims to obtain high diversity level and high convergence speed, all while not suffering from the abovementioned drawbacks of LAHC.
We have termed the proposed method as Diversified Late Acceptance Search (DLAS).
We overview the approach as follows.
We aim to keep or obtain larger fitness values in the fitness array when the search encounters non-improving moves (diversification).
Furthermore, we cautiously replace large fitness values with small values when the search accepts improving moves (intensification).
Lastly, our acceptance criterion is more relaxed than LAHC (diversification).

\subsection{Acceptance Strategy}
\vspace{-1ex}

Comparing the fitness values of the candidate solutions with a larger value than
{\small $\Phi[l]$} (with {\small $l=k \textsf{ mod } L$})
arguably increases diversity of accepted solutions.
Our acceptance strategy is to compare the fitness value {\small $F'$} of the candidate solution {\small $S'$}
in each iteration $k$ with the  maximum fitness value in the fitness array {\small $\Phi$},
instead of comparing it just with {\small $\Phi[l]$}.
The new candidate solution {\small $S'$} would be accepted if {\small $F' = F$} or {\small $F' < \Phi_{\textrm{max}}$},
ie., the maximum value in the fitness array {\small $\Phi$}.
The first condition allows accepting new candidate solutions
with fitness values equal to {\small $\Phi_{\textrm{max}}$} when all the values in {\small $\Phi$} are the same,
especially in the initial and final iterations of the search.
Accepting candidate solutions with smaller fitness values than {\small $\Phi_{\textrm{max}}$}
in other iterations increases the level of acceptable worsening moves
and thereby increases the diversity level of the search.
Section~\ref{DLASsub} shows how to efficiently find and maintain the maximum value in {\small $\Phi$}.

\vspace{-2ex}
\subsection{Replacement Strategy}
\vspace{-1ex}

Our proposed replacement strategy has two parts.
In the first part, if the fitness value {\small $F$} of the new current solution {\small $S$} is larger than {\small $\Phi[l]$},
the value in {\small $\Phi[l]$} is always replaced by {\small $F$}.
Such a replacement is avoided in the most recent version of LAHC to increase intensification of the search.
However, this replacement increases the probability of accepting more worsening moves in future iterations
and thereby can result in better final solutions.
In the second part, if {\small $F$} is smaller than {\small $\Phi[l]$},
the replacement must be done just when {\small $F$} is smaller than the previous value of {\small $F$} as well.
Such a replacement strategy avoids replacing other large values in the fitness array in a series of consecutive steps
if the search falls in a plateau or local optimum.

We note that the combination of the above two replacement approaches is new
and is different from replacing just in acceptance or just in improving moves.
An illustration of the proposed method, especially the replacement strategy, is given in Section~\ref{illustSub}.

\vspace{-2ex}
\subsection{Diversified Late Acceptance Search}
\label{DLASsub}
\vspace{-1ex}
 
\figurename~\ref{algoDLAS} shows the pseudo code for the proposed method using the above acceptance and replacement strategies.
Variables {\small $\Phi_{\textrm{max}}$} and {\small $N$} in \figurename~\ref{algoDLAS} are respectively always equal to the maximum value in the fitness array
and the number of occurrences of that value in the array.
In line 5, {\small $\Phi_{\textrm{max}}$} and {\small $N$} are initialised by {\small $F$} and {\small $L$}.
In every iteration in line 8, {\small $F^-$} holds the previous value of {\small $F$}.
In line 11, new candidate solution {\small $S'$} is accepted if {\small $F'$ = $F$} or {\small $F'<\Phi_{\textrm{max}}$}.
In line 15, if {\small $F > \Phi[l]$}, replacement is made.
Otherwise, in line 17, if {\small $F < \Phi[l]$} and {\small $F<F^-$}, replacement is made.
However, before making the replacement this time, if {\small $\Phi[l]$} is equal to {\small $\Phi_{\textrm{max}}$}, {\small $N$} is decremented by one.
In line 21, if {\small $N$} is zero, the values of {\small $\Phi_{\textrm{max}}$} and {\small $N$} are recomputed by checking all the values in the fitness array.

\begin{figure}[!b]
  \centering
  \vspace{-1ex}
  \includegraphics[width=1.025\textwidth,height=0.5\textwidth]{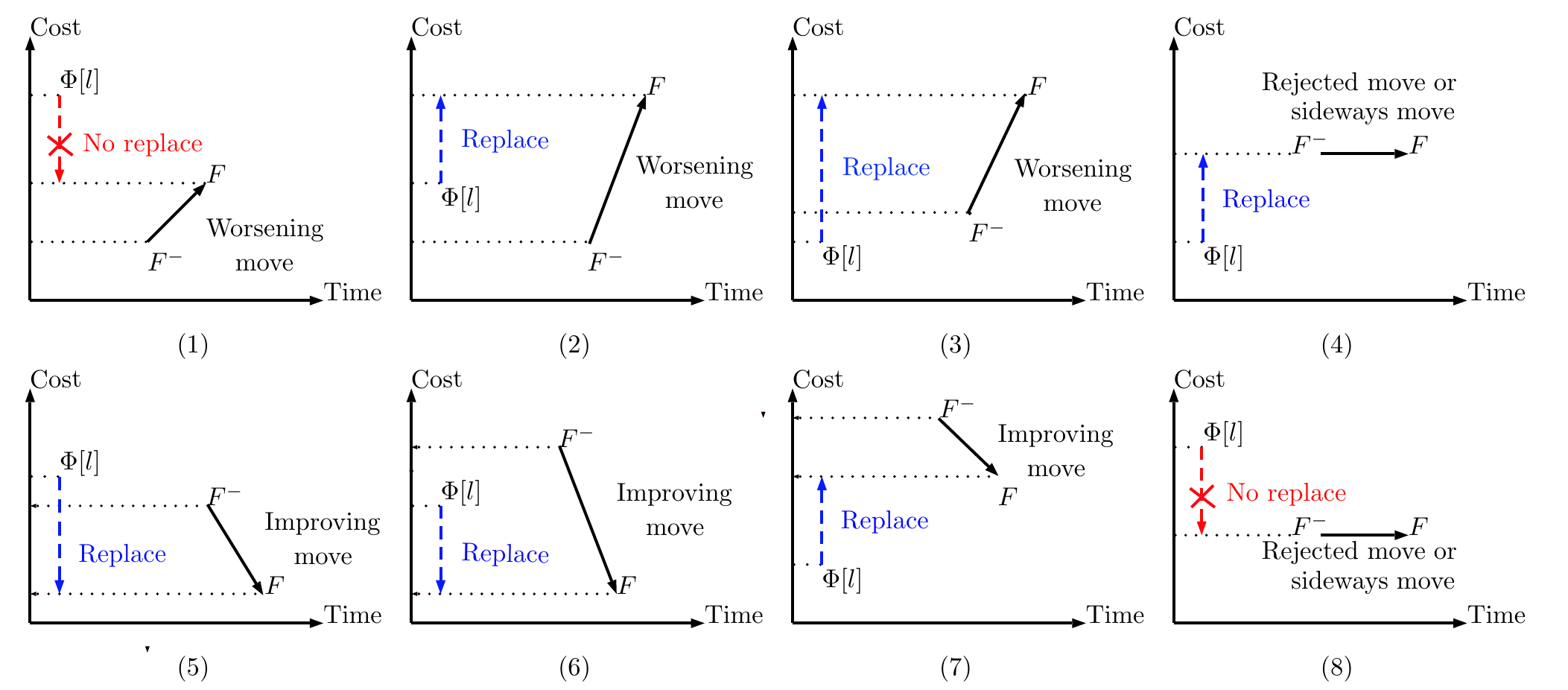}

  \vspace{-2ex}
  \caption
    {
    All possible combinations of values of $F$, $F^-$ and $\Phi[l]$ compared to each other and corresponding replacement rules in the proposed DLAS approach.
    See the text for details.
    }
  \label{DLAS}
\end{figure}

\subsection{DLAS Replacement Scenarios}
\label{illustSub}
\vspace{-1ex}

\figurename~\ref{DLAS} shows eight possible combinations of values of $F$, $F^-$ and $\Phi[l]$ compared to each other and corresponding replacement rules.

\vspace{1ex}
\noindent
{\bf Worsening Moves.}
In cases (1)--(3) in \figurename~\ref{DLAS}, worsening moves take place.
In case~(1), the fitness 
value of the new current solution $F$ is still smaller than $\Phi[l]$. In this case, contrary to 
LAHC, replacement is not allowed in the proposed DLAS method.
This avoidance of replacement actually preserves the large values in the fitness array $\Phi$
when DLAS does not improve the current solutions in some consecutive iterations,
and at the same time the fitness values of the new worse solutions
are not larger than the corresponding values in the fitness array $\Phi$.
In cases (2) and (3), the fitness value of the new current solution $F$ is greater than $\Phi[l]$.
In both these cases, contrary to LAHC, replacement is allowed in DLAS to increase diversity of values in the fitness array $\Phi$. 

\vspace{1ex}
\noindent
{\bf Improving Moves.}
In cases (5)--(7), improving moves take place. In cases (5) and (6),
the fitness value of the new current solution $F$ is smaller than $\Phi[l]$.
In both these cases, as in LAHC, replacement is allowed to optimistically increase the intensification of the search.
In case~(7), the fitness value of the new current solution $F$ is still greater than $\Phi[l]$.
Contrary to LAHC, replacement is allowed in DLAS to increase diversity of values in the fitness array. 

\vspace{1ex}
\noindent
{\bf Sideways Moves or Rejected Moves.}
In cases (4) and (8), there are two possible outcomes:
a candidate solution is not accepted, or a sideways move occurs.
In case (4), the fitness values of the previous and the new current solutions,
ie., $F^-$ and $F$, are greater than $\Phi[l]$.
In this case, contrary to LAHC, replacement is allowed in DLAS to increase diversity of the accepted solutions in future iterations.
In case (8), the fitness values of the previous and the new current solutions are smaller than $\Phi[l]$.
In this case, contrary to LAHC, replacement is not allowed in DLAS. 
This avoidance of replacement actually avoids replacing all the values in the fitness array $\Phi$ 
in consecutive iterations when DLAS falls in a plateau or local optimum.

\vspace{-1ex}
\section{Comparative Evaluation}
\label{secExperiments}
\vspace{-1ex}

In this section we evaluate the performance of the proposed DLAS method,
the most recent version of LAHC (as described in Sec.~\ref{secLateAccept}),
and the recently proposed Step Counting Hill Climbing (SCHC)~\cite{bykov2016stepSCHCUT}.
All experiments were ran on the same computing cluster with a 500~Mb memory limit.
Each node of the cluster is equipped with Intel Xeon CPU E5-2670 processors running at 2.6~GHz.

In SCHC a fitness bound and a counter limit are used instead of a fitness array.
The fitness bound is initialised by the fitness  of the initial solution and the counter limit is similar to the length of the fitness array.
In each iteration, a candidate solution is accepted if its fitness is equal to or better than that of the current 
solution or better than the fitness bound.
Whenever the number of iterations becomes a factor of the counter limit,
the fitness bound is made equal to the fitness of the current solution.

The proposed DLAS algorithm, as well as LAHC and SCHC,
are general purpose local search algorithms for solving any optimisation problem.
Hence, we use sets of Travelling Salesman Problems (TSPs) and Quadratic Assignment Problems (QAPs) 
just to compare the relative performance of the three algorithms,
and not to improve the best known solutions for the individual problems.

\vspace{-2ex}
\subsection{Time Cutoff and Fitness Array Length}
\vspace{-1ex}

To provide a fair comparison, we use time cutoff as the stopping condition.
However, as each instance has its own size and complexity level,
we decided to solve all of them first with LAHC using a reasonably large fitness array size $L$.
We initially performed 50 runs of the LAHC algorithm (with $L$=$50000$) on each instance,
with the stopping condition as getting trapped in a local optimum for at least 10\% of the total running time.
Then we took the longest running time across the 50 runs as the cutoff time for each instance.
We then ran all three algorithms with just the cutoff time as the stopping condition 50 times for each unique value for $L$.

The reported results in the following subsections
are the averages of 50 runs on each instance using the best performing value for $L$. For example,  
\figurename~\ref{U1817Exp} compares LAHC and DLAS algorithms
in the later steps of solving U1817 TSP instance using various values for $L$.
The figure shows that given 290 seconds as the cutoff time for this instance,
$L$=$50000$ and $L$=$5$ are the best values for LAHC and DLAS algorithms, respectively.

\vspace{-2ex}
\subsection{Experiments on TSP instances}
\label{sec:experiments_tsp}
\vspace{-1ex}

Every TSP instance includes a set of cities or points on a map.
The cities are all connected with each other by symmetric roads of given distances or lengths.
The goal of solving such a TSP instance is to find the shortest closed tour that 
includes all the cities such that every city is visited exactly once.
We took all the symmetric Euclidean distance TSP instances
with 1,000 to 10,000 cities from the well-known TSPLIB benchmark dataset at 
{\tt http://comopt.ifi.uni-heidelberg.de/software/TSPLIB95/}.
We used the same source code and the same perturbation heuristic
provided by the authors of~\cite{BURKE201770LAHC17} for solving the TSP instances.
The perturbation heuristic randomly divides a given tour into two parts and then reverses one part~\cite{lin1973effectiveDB}.

\begin{table}[!tb]
\caption
  {
  Results on TSP instances for LAHC and SCHC with {$L$=50000}, and DLAS with {$L$=5}.
  In the first column, the size of each instance is the number in the name of the instance, which indicates the number of cities.
  The 2nd column is the best known solution cost reported in the literature.
  The 3rd column is the time cutoff value used by all methods.
  The 4th column shows the deviations of the best solution cost from the best known solution cost. 
  The 5th column shows the time spent by each algorithm to find the best solution.
  The 6th column shows percentage of iterations in which each algorithm undesirably behaves like HC.
  Shading denotes winning numbers where the differences are statistically significant.
  }
\label{tabTSPtest1}
\centering
\vspace{-2ex}
\scalebox{0.94}
{
\begin{tabular}{|l|r|r|rrr|rrr|rrr|}\hline
 
 &  & &\multicolumn{3}{c|}{\bf Dev. from the best} & \multicolumn{3}{c|}{\bf Time to find the} & \multicolumn{3}{c|}{\bf \% of iterations}\\ 
 
\multicolumn{1}{|c|}{\bf Instance} & \multicolumn{1}{|c|}{\bf Best known}& \multicolumn{1}{|c|}{\bf Time} &\multicolumn{3}{c|}{\bf known solution} & \multicolumn{3}{c|}{\bf last best sol.} & \multicolumn{3}{c|}{\bf behaving like HC }\\\cline{4-12} 

\multicolumn{1}{|c|}{\bf name}	& \multicolumn{1}{|c|}{\bf sol. cost} &\multicolumn{1}{|c|}{\bf cutoff} &{\bf LAHC} & {\bf SCHC}& {\bf DLAS}  & {\bf LAHC} & {\bf SCHC} &{\bf DLAS}  & {\bf LAHC} &{\bf SCHC}& {\bf DLAS} \\\hline\hline

\textbf{Dsj1000} & 18659688 & 100 & 924536 & 705626 & \textbf{339555} & 80 & 66 & 52 & 21 & 36 & \textbf{0} \\ 
\textbf{Pr1002} & 259045 & 120 & 6265 & 6552 & \hl{\textbf{4795}} & 78 & 63 & 51 & 37 & 47 & \textbf{0}  \\ 
\textbf{U1060} & 224094 & 150 & 4560 & 5647 & \hl{\textbf{4193}} & 84 & 68 & 55 & 45 & 54 & \textbf{0} \\ \hline
\textbf{Vm1084} & 239297 & 155 & \textbf{5884} & 6593 & 5927 & 79 & 65 & 51 & 51 & 60 & \textbf{0} \\ 
\textbf{Pcb1173} & 56892 & 160 & 1910 & 2118 & \hl{\textbf{1306}} & 81 & 77 & 49 & 52 & 52 & \textbf{0}  \\ 
\textbf{D1291} & 50801 & 165 & 2612 & 1856 & \textbf{1404} & 111 & 88 & 93 & 35 & 49 & \textbf{0}  \\ \hline
\textbf{Nrw1379} & 56638 & 177 & 2024 & 2159 & \hl{\textbf{1180}} & 117 & 93 & 90 & 37 & 51 & \textbf{0}  \\ 
\textbf{Fl1400} & 20127 & 180 &\hl{\textbf{290}} & 324 & 901 & 116 & 92 & 33 & 43 & 57 & \textbf{0}  \\ 
\textbf{U1432} & 152970 & 200 & 3513 & 4139 &\hl{ \textbf{2022}} & 125 & 114 & 176 & 45 & 55 & \textbf{0}  \\ \hline
\textbf{Fl1577} & 22249 & 250 & \hl{\textbf{466}} & 524 & 634 & 153 & 139 & 108 & 50 & 57 & \textbf{0} \\
\textbf{D1655} & 62128 & 270 & 2424 & 2464 & \hl{\textbf{1550}} & 153 & 120 & 160 & 43 & 59 & \textbf{0}\\ 
\textbf{Vm1748} & 336556 & 280 & 10328 & 11009 & \hl{\textbf{8967}} & 163 & 125 & 173 & 45 & 59 & \textbf{0} \\ \hline
\textbf{U1817} & 57201 & 290 & 2320 & 2461 & \hl{\textbf{1450}} & 189 & 146 & 244 & 41 & 59 & \textbf{0}  \\ 
\textbf{D2103} & 80450 & 309 & 5846 & 6137 & \hl{\textbf{2660}} & 194 & 161 & 279 & 39 & 47 & \textbf{0}  \\ 
\textbf{U2152} & 64253 & 320 & 2598 & 2956 & \hl{\textbf{1350}} & 211 & 198 & 292 & 46 & 51 & \textbf{0} \\ \hline
\textbf{U2319} & 234256 & 350 & 3625 & 3837 & \hl{\textbf{2557}} & 258 & 228 & 347 & 45 & 56 & \textbf{0} \\ 
\textbf{Pr2392} & 378032 & 370 & 19557 & 16025 & \hl{\textbf{9003}} & 238 & 167 & 274 & 40 & 58 & \textbf{0} \\ 
\textbf{Pcb3038} & 137694 & 521 & 6530 & 7118 & \hl{\textbf{3116}} & 324 & 267 & 384 & 42 & 51 & \textbf{0} \\ \hline
\textbf{Fl3795} & 28772 & 1110 & 1542 & 1547 & \hl{\textbf{1202}} & 802 & 769 & 666 & 65 & 72 & \textbf{0} \\ 
\textbf{Fnl4461} & 182566 & 1150 & 9607 & 10558 & \hl{\textbf{3978}} & 454 & 419 & 940 & 62 & 69 & \textbf{0} \\ 
\textbf{Rl5915} & 565530 & 1200 & 36974 & 39929 & \hl{\textbf{19232}} & 718 & 613 & 1198 & 48 & 59 & \textbf{0}  \\ \hline
\textbf{Rl5934} & 556045 & 1320 & 35718 & 38535 & \hl{\textbf{34863}} & 812 & 664 & 814 & 46 & 60 & \textbf{0}\\
\textbf{Pla7397} & 23260728 & 2545 & 962561 & 990251 &\textbf{916947}& 1926 & 1818 & 2542 & 59 & 70 & \textbf{0}\\ \hline
\end{tabular}
}
%\vspace{-1ex}
\end{table}

Table~\ref{tabTSPtest1} shows the results on TSP instances using LAHC and SCHC with $L$=$50000$ and DLAS with $L$=$5$.
The size of each instance is the number in the name of the instance, which indicates the number of cities.
In 20 out of 23 instances, the proposed DLAS method with $L$=$5$ has found better solutions than both LAHC and SCHC with $L$=$50000$.
In 17 of those instances the differences are statistically significant based on t-test with the confidence level of 0.95.
The results also show that in small instances (with small number of cities), DLAS finds better solutions in less time,
while in large instances it does not get trapped in a local optimum quickly and continues to search for a better solution.
For example, for the largest instance in the last line of the table with the time cutoff of 2545 seconds,
LAHC and SCHC are quickly trapped in a local optimum and cannot improve their last found solutions.
In contrast, the proposed DLAS method continues to improve its solutions until almost the end of the cutoff time.

The results also show that even when using a very large value for $L$ in LAHC and SCHC,
in about half of the iterations (especially for large instances), LAHC and SCHC undesirably behave like HC.
This includes iterations in which the maximum value in the fitness array in LAHC
and the fitness bound in SCHC are equal to the last found best solution.
In contrast, the percentage of iterations in which DLAS behaves like HC is zero.
In other words, even when using very small fitness arrays,
there is always room for worsening moves to be accepted by DLAS.
This indicates that the combination of the new acceptance and replacement strategies in DLAS
is more effective in increasing the diversity level of the search than just increasing the length of the fitness array.

\begin{figure}[!t]
\begin{minipage}[t]{1\textwidth}
  \centering
  \begin{minipage}[t]{0.48\textwidth}
    \centering
    \includegraphics[width=1.0\textwidth]{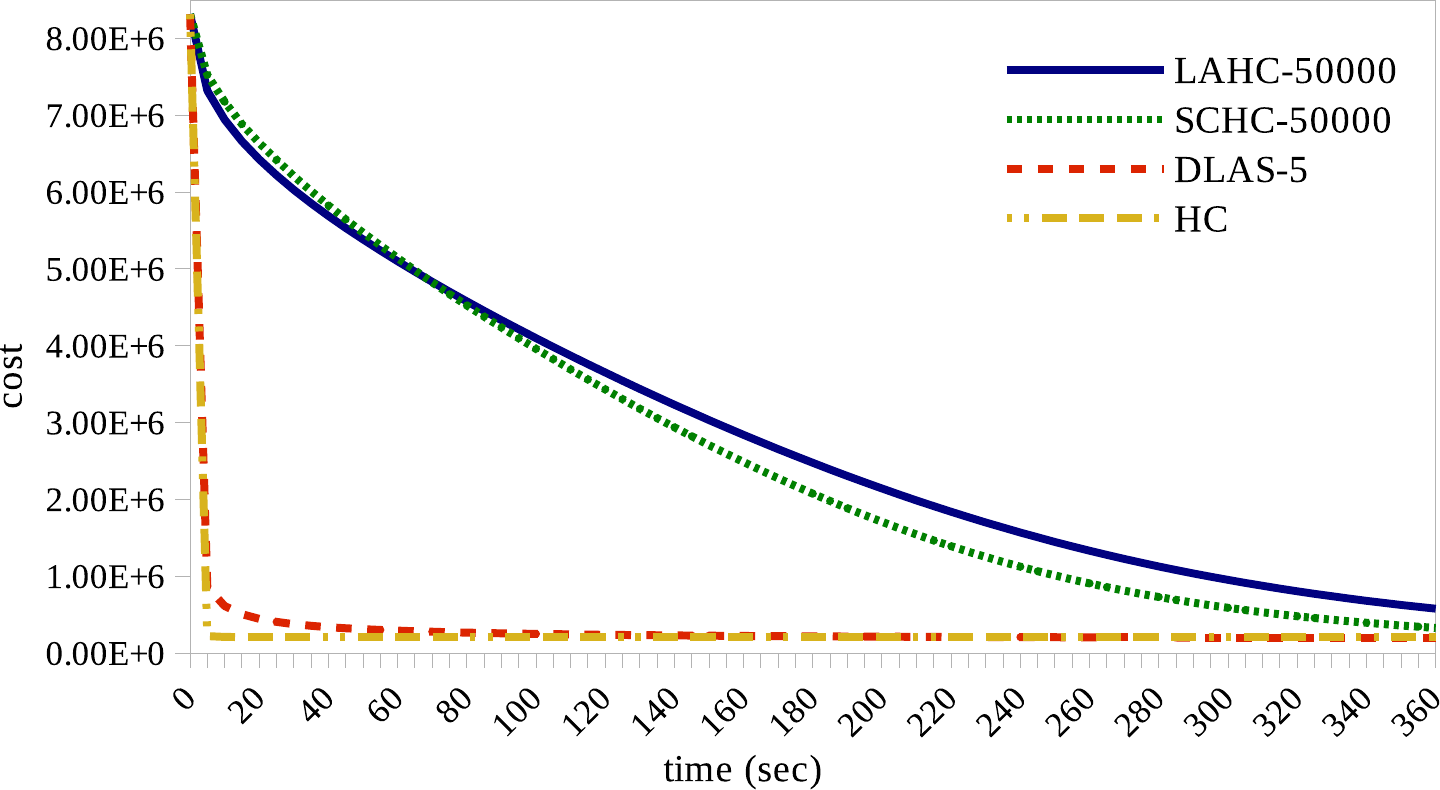}
    \caption
      {
      Search progress for the first 360 seconds while solving the benchmark Fnl4461 TSP instance
      via HC, LAHC and SCHC with {\small $L$}={\small $50000$},
      and DLAS with {\small $L$}={\small $5$}.
      }
    \label{tspIns20a}
  \end{minipage}
  \hfill
  \begin{minipage}[t]{0.48\textwidth}
    \centering
    \includegraphics[width=1.0\textwidth]{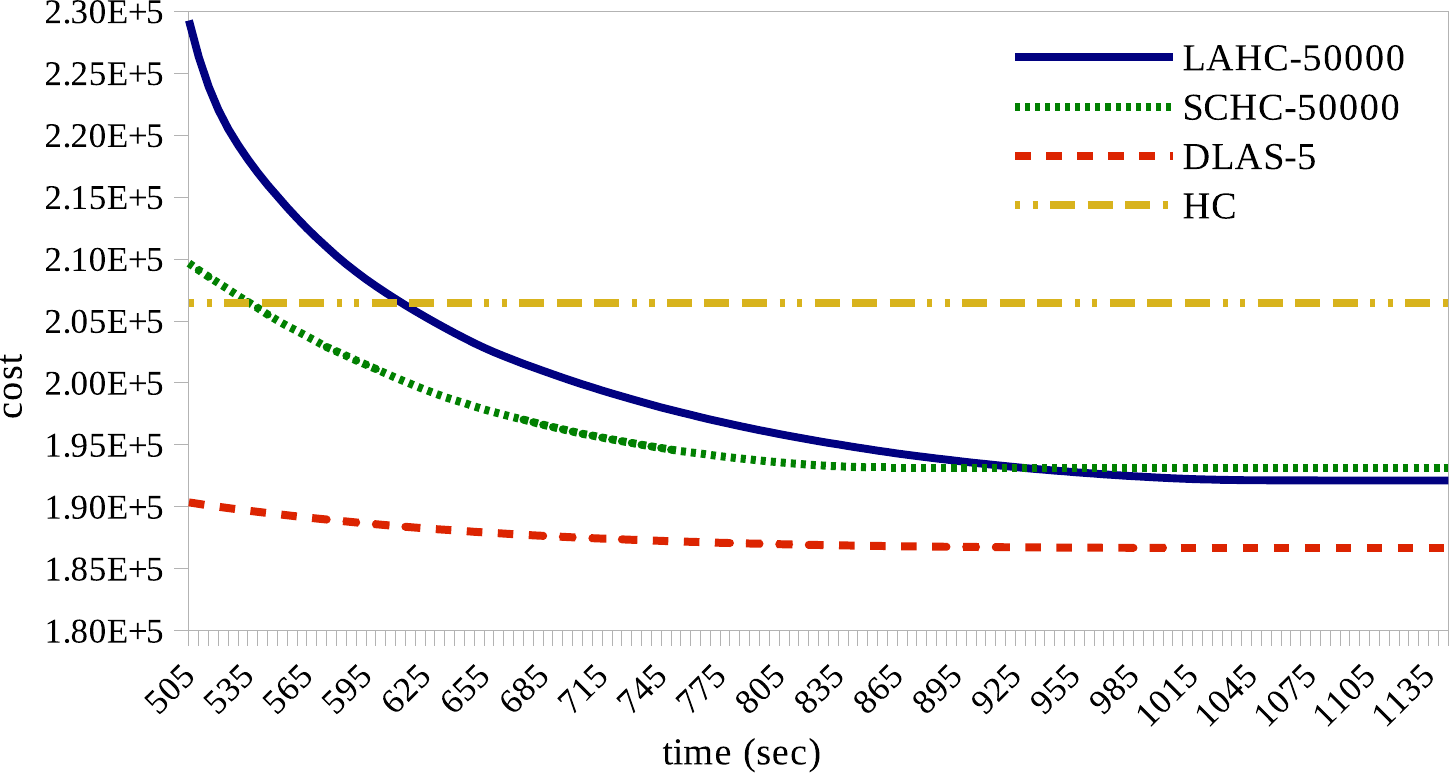}
    \caption
      {
      As per Fig.~\ref{tspIns20a}, but in later iterations.
      The proposed DLAS approach obtains a better solution than HC, LAHC and SCHC.
      Furthermore, DLAS converges quicker than LAHC.
      }
    \label{tspIns20b}
  \end{minipage}
\end{minipage}
\end{figure}

Figs.~\ref{tspIns20a} and~\ref{tspIns20b} show that DLAS with $L$=$5$ has a high convergence speed
(due to the small fitness array size) and converges almost as fast as HC.
It also shows that DLAS with $L$=$5$ ends up with a better solution than LAHC and SCHC with $L$=$50000$,
and HC for the Fnl4461 instance.

\vspace{-2ex}
\subsection{Experiments on QAP instances}
\vspace{-1ex}

Every QAP instance includes two same-size sets of locations and facilities.
The locations are all connected with each other by symmetric links of given distances or lengths.
There is a flow between every pair of facilities with a given weight.
The goal of solving such a QAP instance is assigning each facility to a location
such that the sum of weights of flows between every two facilities 
multiplied by the distances between their assigned locations is minimised.

We took all QAP instances with at least 80 locations and facilities from the 
well-known QAPLIB benchmark dataset at {\tt http://anjos.mgi.polymtl.ca/qaplib/}.
We used the same source code and the same perturbation heuristic provided in 
{\tt http://mistic.heig-vd.ch/taillard/} for solving the QAP instances.
The perturbation heuristic randomly selects two locations and swaps their assigned facilities.

\begin{table}[!tb]
%\vspace{-1ex}
\caption
  {
  Results on QAP instances for LAHC and SCHC with $L$=$50000$, and DLAS with $L$=$10$.
  The size of each instance is the number in the name of the instance, which indicates the number of locations or facilities.
  Explanations for the other columns are as per Table~\ref{tabTSPtest1}.
  }
\label{tabQAPtest1}
\vspace{-2ex}
\centering
\scalebox{0.91}
{
\begin{tabular}{|l|r|r|rrr|rrr|rrr|}\hline

 &  & &\multicolumn{3}{c|}{\bf Dev. from the best} & \multicolumn{3}{c|}{\bf Time to find the} & \multicolumn{3}{c|}{\bf \% of iterations}\\ 
 
\multicolumn{1}{|c|}{\bf Instance} & \multicolumn{1}{|c|}{\bf Best known}& \multicolumn{1}{|c|}{\bf Time} &\multicolumn{3}{c|}{\bf known solution} & \multicolumn{3}{c|}{\bf last best sol.} & \multicolumn{3}{c|}{\bf behaving like HC }\\\cline{4-12} 

\multicolumn{1}{|c|}{\bf name}	& \multicolumn{1}{|c|}{\bf sol. cost} &\multicolumn{1}{|c|}{\bf cutoff} &{\bf LAHC} & {\bf SCHC}& {\bf DLAS}  & {\bf LAHC} & {\bf SCHC} &{\bf DLAS}  & {\bf LAHC} &{\bf SCHC}& {\bf DLAS} \\\hline\hline

\textbf{Lipa80a} & 253195 & 20 & 1607 & 1564 & \hl{\textbf{1411}} & 14 & 11 & 8 & 1.3 & 0.3 & \textbf{0.0} \\ 
\textbf{Tai80a} & 13499184 & 21 & 330957 & 354263 & \hl{\textbf{264177}} & 15 & 12 & 15 & 0.5 & 0.0 & \textbf{0.0} \\ 
\textbf{Lipa80b} & 7763962 & 26 & 39769 & 190699 & \hl{\textbf{0}} & 22 & 17 & 8 & 8.0 & 28.5 & \textbf{0.0} \\  \hline
\textbf{Tai80b} & 818415043 & 27 & 4227835 & \hl{\textbf{3574665}} & 979737 & 20 & 17 & 6 & 8.1 & 16.8 & \textbf{0.0} \\ 
\textbf{Sko81} & 90998 & 24 & 222 & 178 & \textbf{113} & 19 & 16 & 5 & 4.7 & 14.8 & \textbf{0.0} \\ 
\textbf{Lipa90a} & 360630 & 23 & 2045 & 2024 &\hl{\textbf{1893}} & 19 & 15 & 13 & 0.0 & 1.0 & \textbf{0.0} \\ \hline
\textbf{Lipa90b} & 12490441 & 36 & 51015 & 20709 & \hl{\textbf{0}} & 29 & 22 & 11 & 15.0 & 33.2 & \textbf{0.0} \\
\textbf{Dre90} & 1838 & 35 & 1575 & 1615 & \hl{\textbf{1450}} & 16 & 12 & 8 & 0.0 & 6.3 & \textbf{0.0} \\ 
\textbf{Sko90} & 115534 & 28 & 321 & 310 & \textbf{219} & 26 & 21 & 8 & 1.2 & 10.0 & \textbf{0.0} \\ \hline
\textbf{Sko100a} & 152002 & 40 & \textbf{190} & 239 & 218 & 32 & 25 & 11 & 4.6 & 16.8 & \textbf{0.0} \\ 
\textbf{Tai100a} & 21052466 & 35 & 460894 & 486157 & \hl{\textbf{378092}} & 23 & 18 & 29 & 0.0 & 0.9 & \textbf{0.0} \\ 
\textbf{Sko100b} & 153890 & 52 & 175 & 173 & \textbf{160} & 30 & 24 & 10 & 9.3 & 16.0 & \textbf{0.0} \\ \hline
\textbf{Tai100b} & 1185996137 & 55 & \hl{\textbf{2711882}} & 2823207 & 5124004 & 34 & 29 & 13 & 12.6 & 38.3 & \textbf{0.0} \\
\textbf{Sko100c} & 147862 & 42 & 147 & 132 & \textbf{121} & 32 & 26 & 11 & 6.6 & 15.6 & \textbf{0.0} \\ 
\textbf{Sko100d} & 149576 & 42 & \textbf{241} & 246 & 245 & 30 & 24 & 10 & 10.7 & 23.8 & \textbf{0.0} \\ \hline
\textbf{Sko100e} & 149150 & 42 & \textbf{150} & 165 & 156 & 31 & 25 & 10 & 5.8 & 19.7 & \textbf{0.0} \\ 
\textbf{Sko100f} & 149036 & 42 & 237 & 232 & \textbf{204} & 33 & 26 & 11 & 7.7 & 16.9 & \textbf{0.0} \\
\textbf{Wil100} & 273038 & 35 & \hl{\textbf{149}} & 171 & 241 & 32 & 26 & 10 & 2.5 & 12.8 & \textbf{0.0} \\ \hline
\textbf{Dre110} & 2264 & 37 & 2031 & 2057 & \hl{\textbf{1782}} & 25 & 19 & 18 & 1.7 & 4.9 & \textbf{0.0} \\ 
\textbf{Esc128} & 64 & 21 & 0 & 0 & 0 & 6 & 5 & 0.3 & 70.0 & 77.0 & \textbf{0.0} \\ 
\textbf{Dre132} & 2744 & 65 & 2522 & 2543 & \hl{\textbf{2140}} & 39 & 30 & 39 & 4.7 & 10.8 & \textbf{0.0} \\ \hline
\textbf{Tai150b} & 498896643 & 105 & \hl{\textbf{1511339}} & 1669639 & 2641722 & 73 & 61 & 56 & 9.2 & 22.8 & \textbf{0.0} \\ 
\textbf{Tho150} & 8133398 & 130 & 9615 & 9282 & \hl{\textbf{6894}} & 80 & 65 & 79 & 14.1 & 23.8 & \textbf{0.0} \\ 
\textbf{Tai256c} & 44759294 & 60 & \textbf{128527} & 132333 & 134885 & 35 & 27 & 54 & 16.9 & 30.9 & \textbf{0.0} \\ \hline
\end{tabular}
}
%\vspace{-3ex}
\end{table}

Table~\ref{tabQAPtest1} shows the results on QAP instances using 
LAHC and SCHC with $L$=$50000$ and DLAS with $L$=$10$, respectively.
In 15 out of 24 instances, the proposed DLAS method with $L$=$10$ found better solutions than both LAHC and SCHC with $L$=$50000$.
In 10 of those instances the differences are statistically significant based on t-test with the confidence level of 0.95.
Notably, the results also show that in most of the instances, especially small ones, DLAS finds better solutions in considerably less time.
The last column shows that even using a very large value for $L$,
in about 10\% of the iterations LAHC behaves like HC.
For SCHC, it is about 20\%.
In contrast, the percentage of iterations in which DLAS behaves like HC is zero.

\section{Main Findings}
\label{secConclusions}
\vspace{-1ex}

The well-known Late Acceptance Hill Climbing (LAHC) search algorithm strives to escape or avoid local optima by deterministically accepting worsening moves.
LAHC stores fitness values of a predefined number of previous solutions in a fitness array
and compares fitness values of candidate solutions against the least recent element in the array,
rather than simply against the fitness value of the current solution.
The fitness values stored in the array are deterministically replaced as the search progresses.
Unfortunately, the behaviour of LAHC can become similar to that of traditional Hill Climbing search
(ie., getting trapped in a local minimum)
when the same fitness value is stored many times in the fitness array,
particularly when a new local optimum is found.

To address the above issue, we have proposed:
{\bf (i)} a new acceptance strategy which increases diversity of the accepted solutions,
and
{\bf (ii)} a new replacement strategy to improve the diversity of the values in the fitness array by taking worsening, improving, and sideways movement scenarios into account.
These strategies improve the overall diversity of the search.

The proposed Diverse Late Acceptance Search (DLAS) method is shown to outperform the current state-of-the-art LAHC method
on benchmark Travelling Salesman Problems and Quadratic Assignment Problems.
The combination of the new acceptance and replacement strategies in DLAS is more effective in increasing
the diversity of the search than just increasing the length of the fitness array,
and can lead to better quality solutions that are obtained with a lower number of iterations
(ie., less time).

Future avenues of exploration include comparative evaluation of DLAS against other LAHC variants~\cite{abuhamdah2010experimentalLARD},
as well as evaluation on other optimisation problems such as high-school timetabling~\cite{burke2004HighInitVal,fonseca2016lateSfLAHC}.

\bibliographystyle{splncs04}
\bibliography{references}

\end{document}